\documentclass{article}

\usepackage[nonatbib, final]{neurips_2021}
\usepackage[numbers]{natbib}
\usepackage{graphicx, subfigure}

\usepackage[utf8]{inputenc} %
\usepackage[T1]{fontenc}    %
\usepackage{hyperref}       %
\usepackage{url}            %
\usepackage{booktabs}       %
\usepackage{amsfonts}       %
\usepackage{nicefrac}       %
\usepackage{microtype}      %
\usepackage{xcolor}         %
\usepackage[disable]{todonotes}
\usepackage{multirow}

\usepackage[normalem]{ulem} %

\newcommand{\gppnew}[1]{\textcolor{olive}{#1}}
\newcommand{\gppold}[1]{\textcolor{olive}{\sout{#1}}}

\newcommand{\FT}[1]{\textcolor{blue}{#1}}

\newcommand{\FTdel}[1]{\textcolor{blue}{\sout{#1}}}

\newcommand{\iondel}[1]{\textcolor{red}{\sout{#1}}}

\newcommand{\aita}{\texttt{r/AmITheAsshole}}

\renewcommand{\gppnew}[1]{#1}

\renewcommand{\gppold}[1]{} %
\renewcommand{\FT}[1]{#1}
\renewcommand{\FTdel}[1]{}
\renewcommand{\iondel}[1]{}

\makeatletter
\renewcommand{\@noticestring}{%
1st Workshop on Human and Machine Decisions
(WHMD 2021) at NeurIPS 2021.}
\makeatother

\title{Explainable Patterns for Distinction and Prediction of Moral Judgement on Reddit}

\author{%
  Ion Stagkos Efstathiadis%
  \\
  Department of Computing \\
  Imperial College London \\  UK \\
  \texttt{istagkos@outlook.com} 
  \\
  \And
  Guilherme Paulino-Passos\\
  Department of Computing \\
  Imperial College London\\
  UK \\
  \texttt{g.passos18@imperial.ac.uk}
  \AND
  Francesca Toni \\
  Department of Computing \\
  Imperial College London\\
  UK \\
  \texttt{ft@imperial.ac.uk} \\
}
\begin{document}

\maketitle

\begin{abstract}
    The forum \aita\ in Reddit hosts discussion on moral issues \gppnew{based on concrete narratives presented by users}. Existing analysis of the forum focuses on its comments, and does not make the underlying data publicly available. In this paper we build a new dataset of comments and also investigate the classification of the posts in the forum. Further, we identify textual patterns associated with the provocation of moral judgement by posts, with the expression of moral stance in comments, and with the decisions of trained classifiers of posts and comments.

\end{abstract}

\section{Introduction}

This paper focuses on predicting and distinguishing moral judgement using
data from Reddit’s \aita\ subreddit, a forum where users pseudonymously recount 
their confrontational experiences in  posts and then ask commenters for their opinions on whether they were `in the wrong' (i.e. using the subreddit terminology, the `assholes') in their stories. The subreddit includes more than 800K labelled posts and 30M labelled comments, thus providing a unique window into the rendering of moral judgement in the anonymous global setting of social media. 

\begin{figure}[h]
    \centering
    \includegraphics[width=0.8\textwidth]{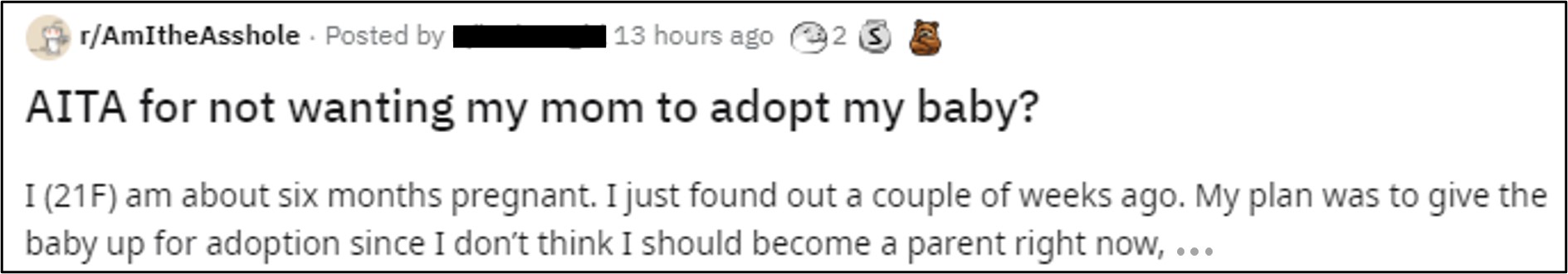}
    \includegraphics[width=0.8\textwidth]{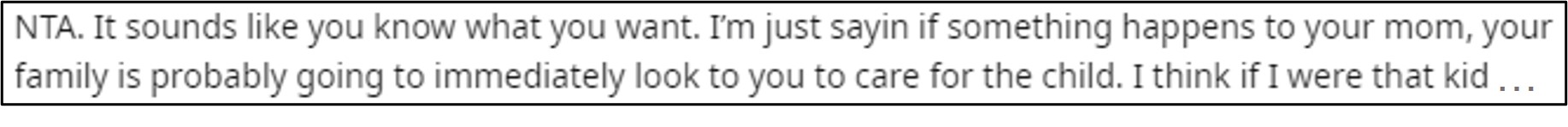}
    \caption{Example of a post excerpt (above) and a comment excerpt (below) in the \aita\ subreddit. The NTA prefix indicates that the commenter judges the poster as “Not the Asshole”.}
    \label{fig:post-comment-eg}
\end{figure}

Example excerpts from a post and a comment addressing it are shown in \autoref{fig:post-comment-eg}. Each comment is instructed to begin with one of four pre-set tags revealing its verdict: \textbf{i)} NTA (Not The Asshole); \textbf{ii)} YTA (You are The Asshole); \textbf{iii)} NAH (No Assholes Here); \textbf{iv)} ESH (Everyone Sucks Here). 18 days after its creation, each post is officially labelled by the tag corresponding to its top comment. Hence, comments are labelled by their composers with the verdicts that they advocate, while posts inherit their verdict labels from their most popular comments.

Existing work on \aita\ \citep{DBLP:journals/corr/abs-2101-07664} develops classifiers from the comments (but not from the posts), achieving  best accuracy of 89\% by fine-tuning BERT \cite{devlin2018bert}, and then explores out-of-distribution uses of the classifiers. Instead, in this paper, we classify posts as well as comments, and explore patterns emerging from the data and from the classifiers.

In summary, our contributions in this paper are as follows. Firstly, we support the prediction of moral judgement in Reddit by predicting the verdicts assigned to posts, achieving best accuracy of 62\% using BERT. This low accuracy compared with that in classifying the comments, highlights the challenges of the task. Secondly,  we construct a novel dataset of comments\footnote{available at \url{https://github.com/CLArg-group/moral-judgement-reddit-whmd-2021}} and build a classifier of moral stance in Reddit by predicting the verdicts that comments advocate, achieving   86\% accuracy using BERT. Thirdly, 
we apply GrASP \cite{shnarch2017grasp, lertvittayakumjorn2021grasp}, to identify textual patterns associated with (i) the provocation and expression of moral judgement and (ii) the decisions of trained moral judgement classifiers. The first type of patterns have sociological value by documenting wordings that are judged in the context of social media. The second renders the decisions of classifiers more transparent, thus encouraging trust in them \cite{giraud1998beyond} and potentially facilitating their debugging \cite{lertvittayakumjorn2020find}.

\section{Related Work}

The classification of \emph{comments} of the \aita\ subreddit has recently been achieved in \citep{DBLP:journals/corr/abs-2101-07664}, with best accuracy of 89\% using \FT{(fine-tuned)} BERT with dropout and a fully connected layer on top.
\FT{We adopt the same approach for classifying comments, but fine-tuning on our novel dataset of comments, and achieving a similar accuracy.} 

In general, stance classification has been largely ignited by \citep{mohammad2016semeval}, which has inspired classifiers of two families: feature-engineering-based as in \cite{hacohen2017stance, kuccuk2018stance, dey2017twitter, sen2018stance, pavan2020morality}, and word-embedding based as in \cite{zarrella2016mitre, wei2016pkudblab, siddiqua2019tweet, ghosh2019stance}.  \FT{Our approach belongs to the second category, as our models are based on BERT.}

Work involving judgement prediction is found in legal prediction, where the outcome of legal trials is predicted based on court judgements, i.e. documents which contain facts about cases. Several works \citep{aletras2016predicting, liu2018two, shaikh2020predicting, bertalan2020predicting} use BoW-based (Bag-of-Words) classifiers to predict the decisions of various courts.%
These works can leverage on the standardised formulation of case facts and consistent labels by courts, possibly including the implicit/explicit judgment \citep{aletras2016predicting}. 
Reddit posts instead lack \gppnew{strict standards}.%

\section{Datasets}

\begin{figure}[hbt]
    \centering
    \includegraphics[width=0.7\textwidth]{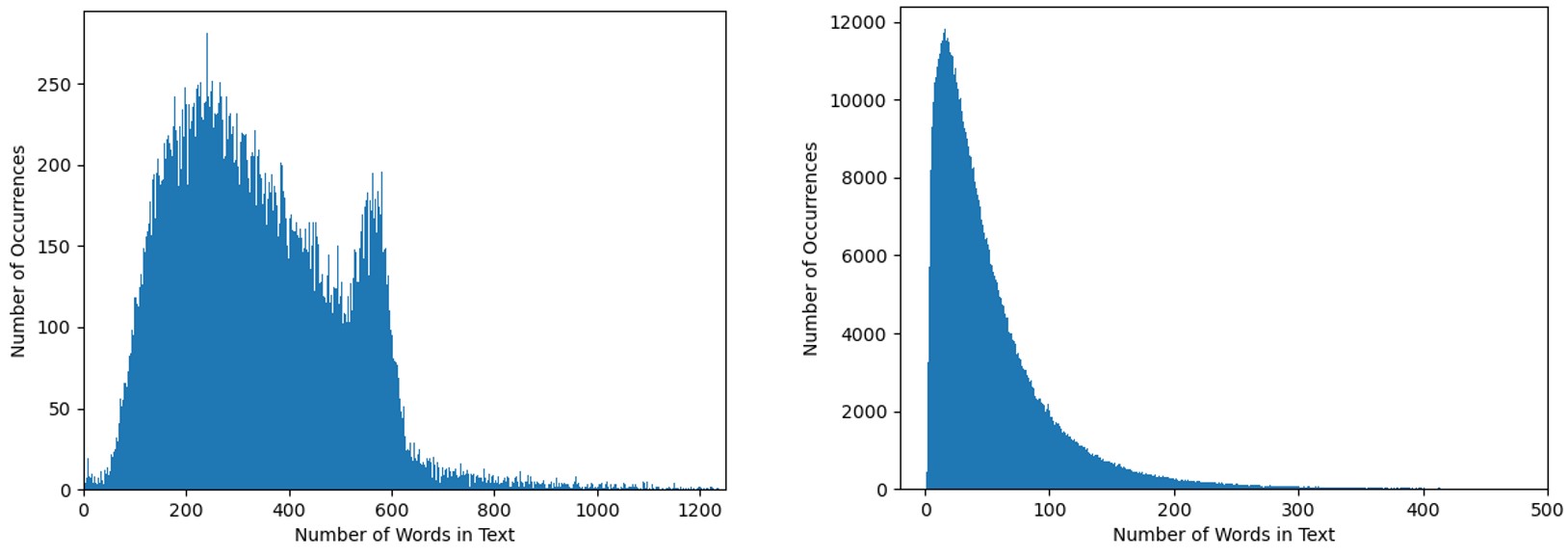}
    \caption{The distribution of text lengths in posts (left) and comments (right).}
    \label{fig:text-lengths}
\end{figure}

\paragraph{Posts.} We download 97K post data from the repository published in \citep{aita2020}. Each datum consists of a text (post title + body) and a label, derived from the post's verdict as follows: NTA and NAH map to 0 for sweethearts (i.e. non-assholes); YTA and ESH map to 1 for assholes. %
\gppnew{\autoref{fig:text-lengths} shows post length distribution, with a considerable number of long posts (over 500 tokens).} %
Further inspection reveals a class imbalance: around 75\% sweethearts and 25\% assholes.

\paragraph{Comments.} We use the Pushshift API \citep{baumgartner2020pushshift} to scrape around 30M comment data from the beginning of \aita\ until the end of 2020.\footnotemark[\value{footnote}] We then discard deleted comments and comments rated lower than 3 for quality control. We use regular expressions to identify the verdicts of comments (by pinpointing YTA, ESH, NAH, NTA), we store the verdicts separately, we discard comments without verdicts, and erase the tags to avoid giving the verdicts away. We finally only keep primary comments to be left with around 600K data. Each datum consists of a comment body (i.e. a text) and a binary label giving the verdict. \autoref{fig:text-lengths} shows that comments are \gppnew{generally} shorter than posts. Further inspection reveals around 66\% sweetheart advocates and 33\% asshole advocates. %

\section{Classifying the Posts}

\paragraph{Classification with BERT.} For classifying the posts, we employ the \texttt{bert-base-cased} model from the \textit{Hugging Face} library \citep{wolf2019huggingface}, with its built-in pre-processing of texts
(implementation hyperparameters are shown in %
the Appendix). \gppnew{Importantly, many posts are longer than 512 words, the maximum sequence length handled by BERT, and thus are truncated.}  %

The results are shown in Table~\ref{table:post-class-results}. The task is proven challenging with the classifier unable to outperform the baseline method of always predicting the majority class. Consequently, to ensure that performance is not boosted by the classifier's favouritism towards sweethearts, we also report results on a rebalanced dataset. An accuracy of 62\% is reported on both test sets. This is higher than the accuracies we achieved with simpler classifiers such as the Multinomial Naive Bayes and LogReg. We believe that further hyperparameter tuning and longer training can further increase it. Moreover, the F1 scores on the rebalanced dataset reveal equitable handling of the two classes. 

\begin{table}[ht]
\caption{Results on test and rebalanced test datasets for the classification of post texts.} %
\label{table:post-class-results}
\centering
\begin{tabular}{c|cccc|cccc}
 & \multicolumn{4}{c|}{Unbalanced test set} & \multicolumn{4}{c}{Rebalanced test set} \\ [0.5ex] \hline
Class & Precision & Recall & F1 & Accuracy & Precision & Recall & F1 & Accuracy \\ [0.5ex]
\hline
assholes & $0.38$ & $0.62$ & $0.47$ & \multirow{2}{*}{$0.62$} & $0.59$ & $0.62$ & $0.61$ & \multirow{2}{*}{$0.62$} \\
sweethearts & $0.81$ & $0.62$ & $0.70$ &  & $0.64$ & $0.61$ & $0.63$ &  \\
[1ex]
\hline
\end{tabular}
\end{table}

\paragraph{Qualitative Analysis with GrASP.} We identify textual patterns associating posts with their actual or predicted labels. We do this using GrASP \citep{shnarch2017grasp}, an algorithm for extracting rich patterns from textual data. Specifically, we use the GrASP library from \cite{lertvittayakumjorn2021grasp}. Examples of the outputs of GrASP for an input post are shown in \autoref{fig:grasp-ex-small} (more examples are given in the Appendix). Here, the score for the resulting patterns is given by their information gain, reflecting their classification value.

\begin{figure}[hbt]
    \centering
    \includegraphics[width=0.8\textwidth]{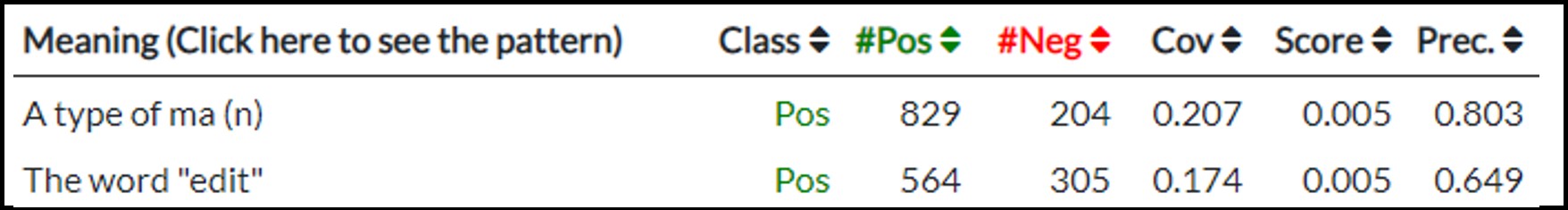}
    \caption{Patterns identified by 
    GrASP to associate posts %
    with their actual
    labels. GrASP is applied on random %
    5000 posts. %
    `Pos'/`Neg' stand for %
    sweethearts/%
    assholes. `Score' refers to information gain.}
    \label{fig:grasp-ex-small}
\end{figure}

In patterns associating posts with their actual labels%
, the highest information gain by a pattern is 0.005 (as in Figure~\ref{fig:grasp-ex-small}), a very small value compared to gains of 0.06 identified in use cases of GrASP in \cite{lertvittayakumjorn2021grasp}. Such small information gains are in line with the difficulty of BERT in classifying the posts. %
Also,
note that GrASP blindly associates each pattern with the class that contains it the most but this may be misleading in cases of class imbalance. For example, in Figure~\ref{fig:grasp-ex-small}, even though the word ``edit'' is reported as indicative of sweetheart, a text containing the word ``edit'' is more likely to be asshole than one picked at random from the 5000 samples.%

In patterns associating posts with their labels as predicted by BERT (see the Appendix for some examples), information gains are significantly larger, suggesting that classes assigned by BERT are more consistent than actual classes. Reassuringly, BERT patterns generally agree with actual patterns. Notably, BERT did not pick up on the word ``edit'' being associated with assholes, perhaps because in its pre-training, BERT did not associate a polarity with this word, which in other contexts is more neutral, and thus very far from other class indicative words.

\section{Classifying comments} 

\paragraph{Classification with BERT.} We employ the same BERT model as for the posts (implementation hyperparameters are shown in %
the Appendix). Results are shown in \autoref{table:comment-class-results}. %

\begin{table}[ht]
\centering
\caption{Results on test dataset for the classification of comments. Accuracy is over the entire test set.}
\label{table:comment-class-results}
\begin{tabular}{c|cccc}
\hline
Class & Precision & Recall & F1 & Accuracy\\ [0.5ex]
\hline
assholes & $0.78$ & $0.83$ & $0.80$ & \multirow{2}{*}{$0.86$} \\
sweethearts & $0.91$ & $0.88$ & $0.90$ & \\
\hline
\end{tabular}
\end{table}

Here, the classifier outperforms the baseline method of always predicting the majority class. Thus, results are only reported on the unbalanced test set representative of class proportions in the population. We report an accuracy of 86\%, which verifies that comments are easier to classify than posts. Our accuracy is slightly lower than the 89\% reported in \citep{DBLP:journals/corr/abs-2101-07664} possibly due to a differently constructed dataset (not publicly available). We believe that further hyperparameter tuning and longer training can further increase it. The bias conveyed by the deviation in F1 scores with respect to the two classes %
may be the result of the imbalance in the test set%
.

\paragraph{Qualitative analysis with GrASP.} Some top patterns associating comments with their actual labels are shown in Figure~\ref{fig:grasp-comments-ex-small} (for more patterns see Figure \ref{fig:grasp-comments} (a) in the Appendix). In general, patterns drawn from the comments offer slightly higher information gains than patterns drawn from the posts, which agrees with the classification of comments being easier. There is also greater balance in patterns associated with the two classes, which benefits the classification of the comments.

\begin{figure}[hbt]
    \centering
    \includegraphics[width=0.8\textwidth]{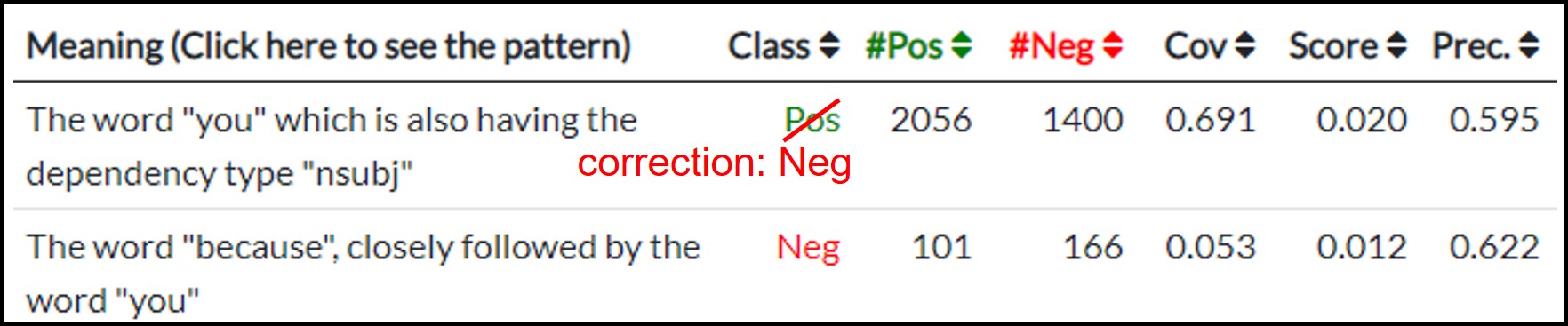}
    \caption{Patterns identified by 
    GrASP to associate comments 
    with their actual
    labels. GrASP is applied on random %
    5000 comments. %
sweethearts/ 
    assholes. `Score' refers to information gain. Manual corrections are applied on patterns that have been associated with sweetheart although they are arguably indicative of asshole.}
    \label{fig:grasp-comments-ex-small}
\end{figure}

Examples of patterns associating comments with their  labels as predicted by BERT can be found in the \ref{fig:grasp-comments} (b) of the Appendix. Reassuringly, upon closer inspection, patterns are found to agree with those associating comments with their actual labels. Notably, as with BERT patterns for posts, the information gains of patterns spotted by GrASP to associate comments with their BERT-predicted labels, are significantly higher than gains from the patterns of actual labels. %

\section{Conclusions}

We show that the prediction of moral judgement by classifying the posts of the \aita\ subreddit is challenging, with the best accuracy of 62\% achieved using BERT. The main challenges of the task are the following: \textbf{i)} long texts exceeding the limit of 512 tokens; \textbf{ii)} non-obvious classification requiring deductive thought by humans; \textbf{iii)} noisy labels coming from subjective views of different voters; \textbf{iv)} class imbalance; \textbf{v)} non-standard texts written by different users. For the classification of moral stance in Reddit comments, we construct a new dataset of ~600K data and achieve an accuracy of 86\% using BERT.  We identify textual patterns associated with actual and predicted Reddit moral judgement in posts and comments using GrASP. %

The most obvious improvement for classifications is to use better computational resources to: \textbf{i)} conduct more exhaustive hyperparameter search; \textbf{ii)} completely unfreeze BERT. The noisiness of post labels due to subjectivity could be internalised by assigning each post an \textit{assholeness score} based on its top comments and predicting it in a regression task. Techniques for robust training with noisy data presented in \citep{song2020learning} or multi-task learning techniques to leverage the information in the comments could improve the classification of the posts. An improvement in pattern identification (as in GrASP) could be to augment texts with additional custom attributes, like their pertinence to the moral foundations dictionary \citep{hopp2021extended}. Lastly, we could explain BERT decisions with other techniques such as dictionary learning \citep{yun2021transformer}.

\bibliographystyle{plainnat}
\bibliography{aita}

\newpage
\appendix

\section{Appendix}

\begin{table}[ht]
\caption{Important hyperparameters for the classification of post texts using unfrozen BERT's pooler output.}
\centering
\begin{tabular}{c c c}
\hline\hline
Implementation Part & Hyperparameter & Value \\ [0.5ex]
\hline
data processing & train/valid/test proportions & 0.8/0.1/0.1 \\
data processing & text truncation length & 512 \\
data processing & rebalancing method & reweighting \\
model architecture & BERT output used & pooler \\
model architecture & num unfrozen layers & 10 + pooler layer \\
model architecture & output layer type & fully connected \\
model architecture & dropout & NA \\
model architecture & batch normalisation & NA \\
training & batch size & 8 \\
training & learning rate & 0.0000003 \\
training & optimiser & AdamW \\
training & weight decay & 0.15 \\
training & num training epochs & 6 \\
\hline
\end{tabular}
\label{table:post-class-params}
\end{table}

\begin{figure}[h]
    \centering
    \subfigure[]{\includegraphics[width=0.8\textwidth]{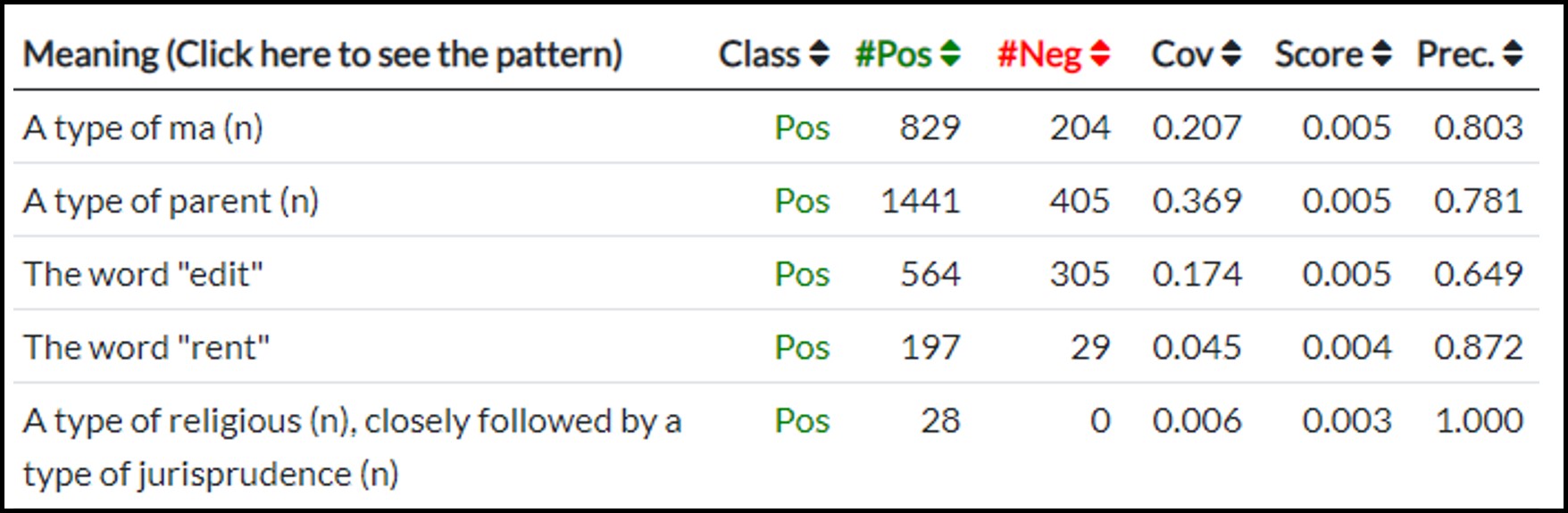}} 
    \subfigure[]{\includegraphics[width=0.8\textwidth]{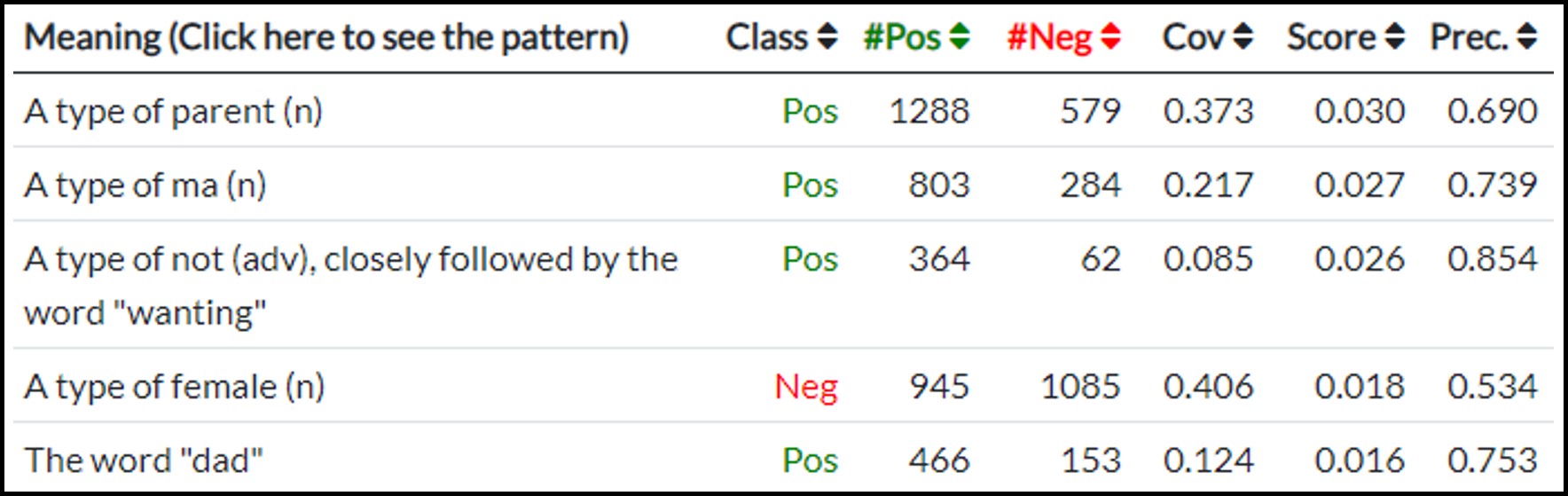}}
    \caption{Patterns identified by GrASP to associate post texts with their (a) actual and (b) predicted labels. GrASP is applied on random samples of 5000 posts. Class `Pos' is that of sweethearts. Class `Neg' is that of assholes. `Score' refers to information gain.}
    \label{fig:grasp-posts}
\end{figure}

\begin{figure}[hbt]
    \centering
    \includegraphics[width=0.8\textwidth]{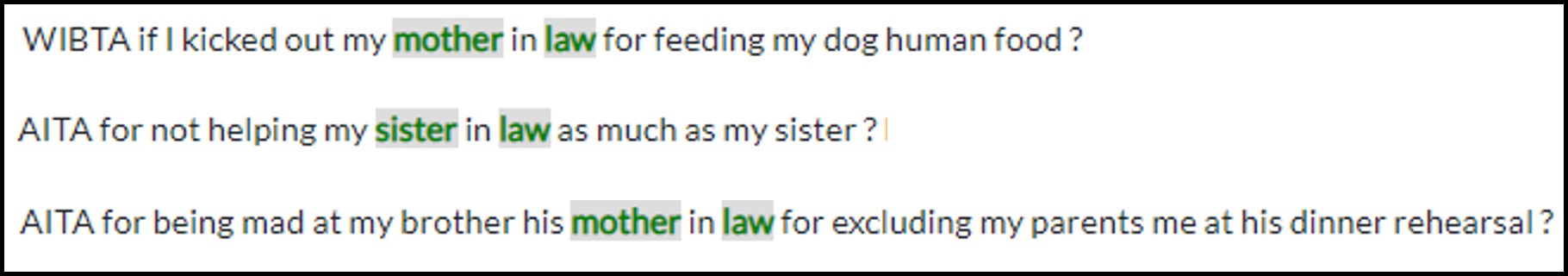}
    \caption{Three examples of insightful excerpts from posts containing the pattern described by GrASP as “a type of religious, closely followed by a type of jurisprudence” (see last pattern in Figure \ref{fig:grasp-posts} (a)).}
    \label{fig:grasp-inlaws}
\end{figure}

\begin{table}[ht]
\caption{Important hyperparameters for the classification of comment bodies using unfrozen BERT's pooler output.}
\centering
\begin{tabular}{c c c}
\hline\hline
Implementation Part & Hyperparameter & Value \\ [0.5ex]
\hline
data processing & train/valid/test proportions & 0.8/0.1/0.1 \\
data processing & text truncation length & 80 \\
data processing & rebalancing method & reweighting \\
model architecture & BERT output used & pooler \\
model architecture & num unfrozen layers & 8 + pooler layer \\
model architecture & output layer type & fully connected \\
model architecture & dropout & NA \\
model architecture & batch normalisation & NA \\
training & batch size & 16 \\
training & learning rate & 0.0000008 \\
training & optimiser & AdamW \\
training & weight decay & 0.1 \\
training & num training epochs & 6 \\
\hline
\end{tabular}
\label{table:comment-class-params}
\end{table}

\begin{figure}[p]
    \centering
    \subfigure[]{\includegraphics[width=0.8\textwidth]{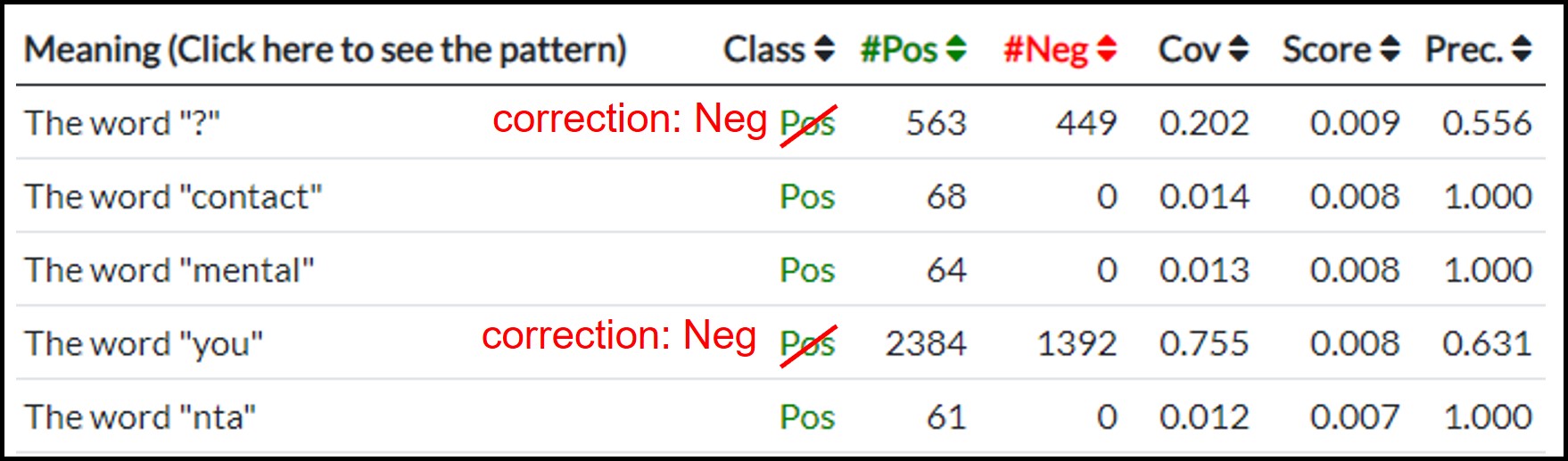}} 
    \subfigure[]{\includegraphics[width=0.8\textwidth]{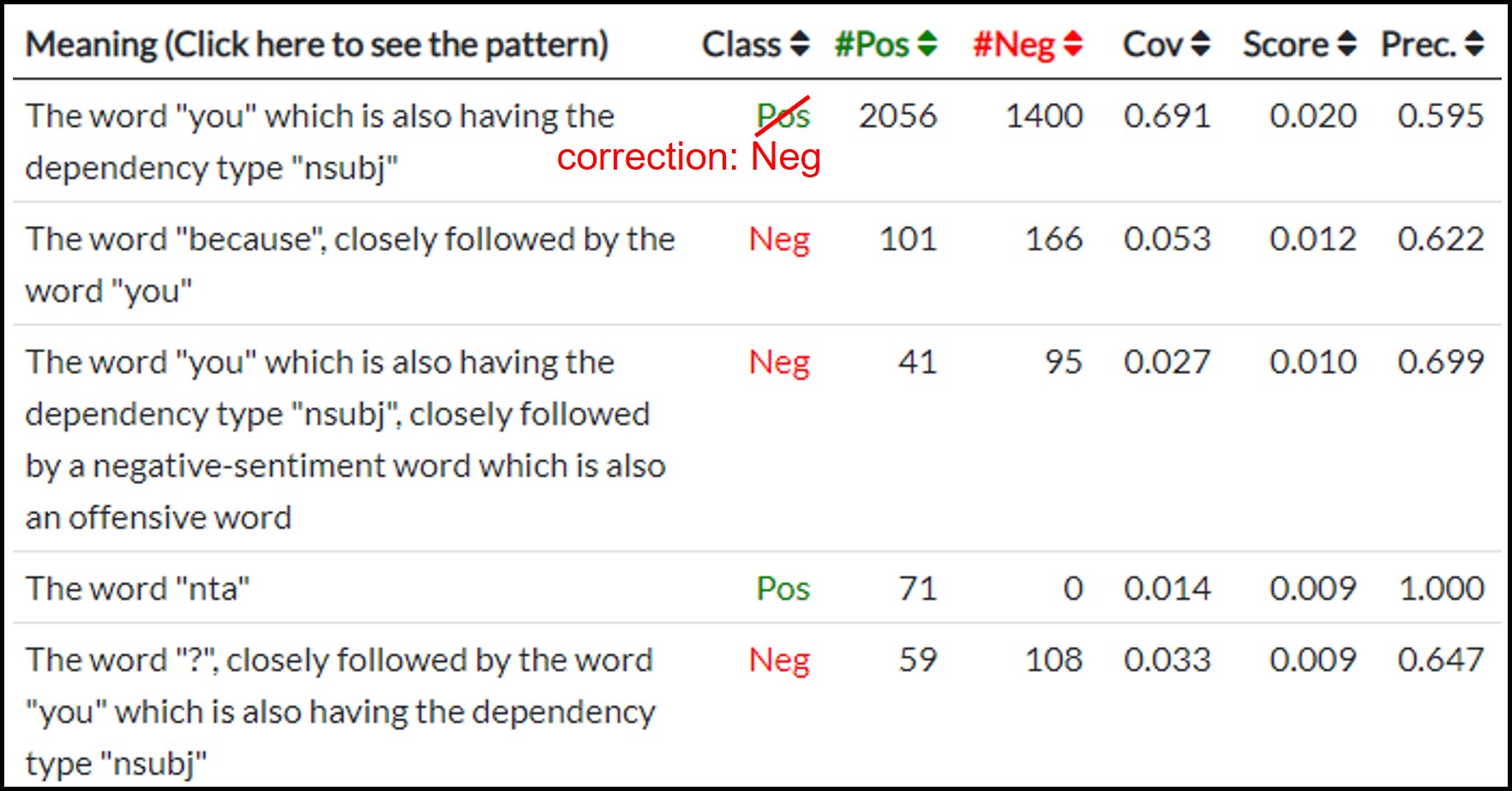}}
    \caption{Patterns identified by GrASP to associate comment bodies with their (a) actual and (b) predicted labels. GrASP is applied on random samples of 5000 comments. Class ‘Pos’ is that of sweethearts. Class‘Neg’ is that of assholes. ‘Score’ refers to information gain. Manual corrections are applied on patterns that have been associated with sweetheart although they are arguably indicative of asshole.}
    \label{fig:grasp-comments}
\end{figure}

\end{document}